\documentclass[letterpaper, 10 pt, journal, twoside]{ieeetran}
\IEEEoverridecommandlockouts % This command is only needed if you want to use the \thanks command

\usepackage[utf8]{inputenc} %latin1
\usepackage{amsmath}
\usepackage{amsfonts}
\usepackage{amssymb}
\usepackage{mathtools}
\usepackage{graphicx}
\usepackage{color}
\usepackage{booktabs}
\usepackage{color}
\usepackage{tabularx}
\usepackage{subcaption} 
\usepackage[bottom]{footmisc}
\DeclareUnicodeCharacter{2212}{-}
\usepackage{pgfplots}
\pgfplotsset{compat=newest}
\usepgfplotslibrary{groupplots}
\usepackage{grffile}
\pgfplotsset{plot coordinates/math parser=false}
\newlength\fheight
\newlength\fwidth

\usepackage{algorithm}
\usepackage{algorithmicx}
\usepackage{algpseudocode}
\algnewcommand\algorithmicAnd{\textbf{and} }
\usepackage{url} % for bibtex

\let\vec\boldvec
\def\code#1{\texttt{#1}} % coding font
\def\CC{{C\nolinebreak[4]\hspace{-.05em}\raisebox{.4ex}{\tiny\bf ++}}}
\newcommand{\joint}{\vec{q}} % used to denote robot state in joint space
\newcommand{\joints}{\vec{Q}} % multiple demonstrations in joint space 
\newcommand{\alg}{\textit{LSDP}}
\newcommand{\algCoupled}{\textit{cLSDP}}
\newcommand{\lone}{\lambda_1}
\newcommand{\ltwo}{\lambda_2}
\newcommand{\numTimes}{N}
\newcommand{\numMovement}{n}
\newcommand{\numDemos}{d}
\newcommand{\numRbf}{p}
\newcommand{\centersRbf}{\vec{\mu}}
\newcommand{\widthsRbf}{\vec{\sigma}^{2}}
\newcommand{\timePoints}{\vec{t}}
\newcommand{\paramsRegr}{\vec{\theta}} % regression parameters
\newcommand{\paramsRbf}{\vec{\beta}} % basis function parameters
\newcommand{\basis}{\vec{\Psi}} % basis functions
\newcommand{\basisAcc}{\ddot{\vec{\Psi}}} % second derivative of basis funcs

\newcommand{\jointMax}{\joint_{\mathrm{max}}}
\newcommand{\jointMin}{\joint_{\mathrm{min}}}

\newcommand{\racketRadius}{r_{R}} % racket radius
\graphicspath{{./}}
\mathtoolsset{showonlyrefs} 

\author{Okan Ko\c c$^{1}$, Jan Peters$^{1,2}$% <-this % stops a space
\thanks{Manuscript received: September, 11, 2018; Revised: December, 8, 2018; Accepted: January, 14, 2019.}
\thanks{This paper was recommended for publication by Editor Dongheui Lee upon evaluation of the Associate Editor and Reviewers' comments.}
\thanks{$^{1}$Okan Ko\c c and Jan Peters are with Max Planck Institute for Intelligent Systems,
	    Max-Planck-Ring 4, 72076 T\"ubingen, Germany
        {\tt\footnotesize okan.koc@tuebingen.mpg.de}}
\thanks{$^{2}$Jan Peters is with Technische Universitaet Darmstadt, FG Intelligente Autonome Systeme
        Hochschulstr. 10, 64289 Darmstadt, Germany {\tt\footnotesize peters@ias.tu-darmstadt.de}}
      \thanks{Digital Object Identifier (DOI): see top of this page.}
}
\title{Learning to Serve: an Experimental Study for a new Learning from Demonstrations Framework}

\begin{document}

\maketitle
% comment following for final version
% \thispagestyle{empty}
% \pagestyle{empty}

%%%%%%%%%%%%%%%%%%%%%%%%%%%%%%%%%%%%%%%%%%%%%%%%%%%%%%%%%%%%%%%%%%%%%%%%%%%
\begin{abstract}

%We evaluate in our table tennis setup the performance of this new
%strategy and show that it compares favorably with a previous
%trajectory generation approach.

  Learning from demonstrations is an easy and intuitive way to show
  examples of successful behavior to a robot. However, the fact that
  humans optimize or take advantage of their body and not of the
  robot, usually called the \emph{embodiment problem} in robotics,
  often prevents industrial robots from executing the task in a
  straightforward way. The shown movements often do not or cannot
  utilize the degrees of freedom of the robot efficiently, and
  moreover can suffer from excessive execution errors. In this paper,
  we explore a variety of solutions that address these
  shortcomings. In particular, we learn sparse movement primitive
  parameters from several demonstrations of a successful table tennis
  serve. The number of parameters learned using our procedure is
  independent of the degrees of freedom of the robot. Moreover, they
  can be ranked according to their importance in the regression
  task. Learning few parameters that are ranked is a desirable feature
  to combat the curse of dimensionality in Reinforcement
  Learning. Real robot experiments on the Barrett WAM for
  a table tennis serve using the learned movement primitives show that
  the representation can capture successfully the style of the
  movement with few parameters.

% optimize these trajectories online,
% correcting for execution and task errors. Moreover, we compare such
% Model Predictive Control (MPC) based approaches to episodic
% Reinforcement Learning (RL). We implement a hybrid model-based
% approach that models the ball from very few examples and performs
% model-free rollouts on the robot, while simulating the outcome using
% an internal ball-model. We find that this hybrid RL approach performs
% quite well in practice. Our results show that both methods can serve
% balls successfully from different initial conditions with a high
% percentage.

\end{abstract}


%%%%%%%%%%%%%%%%%%%%%%%%%%%%%%%%%%%%%%%%%%%%%%%%%%%%%%%%%%%%%%%%%%%%%%% 
%%%%

\section{Introduction}

\IEEEPARstart{H}{umans} are good at using their bodies to great effect, taking
advantage of their muscular structure and soft but flexible
actuation. Much of dexterous manipulation, or dynamic movement
generation reflects this awareness of the human body. When teaching
the robots to achieve similar tasks autonomously, however, we
inevitably impose and transfer our biases to the robot. This problem
of \emph{embodiment} can cripple the execution, possibly also preventing
the robots from taking advantage of their kinematics structure and
actuation mechanisms.

% mention execution errors or torque-mismatch as one of the primary
% reasons why we can not transfer successful behaviour so easily.

\begin{figure*}[t]
  \centering\tiny
  \begin{minipage}{.25\textwidth}
    \includegraphics[scale=0.10]{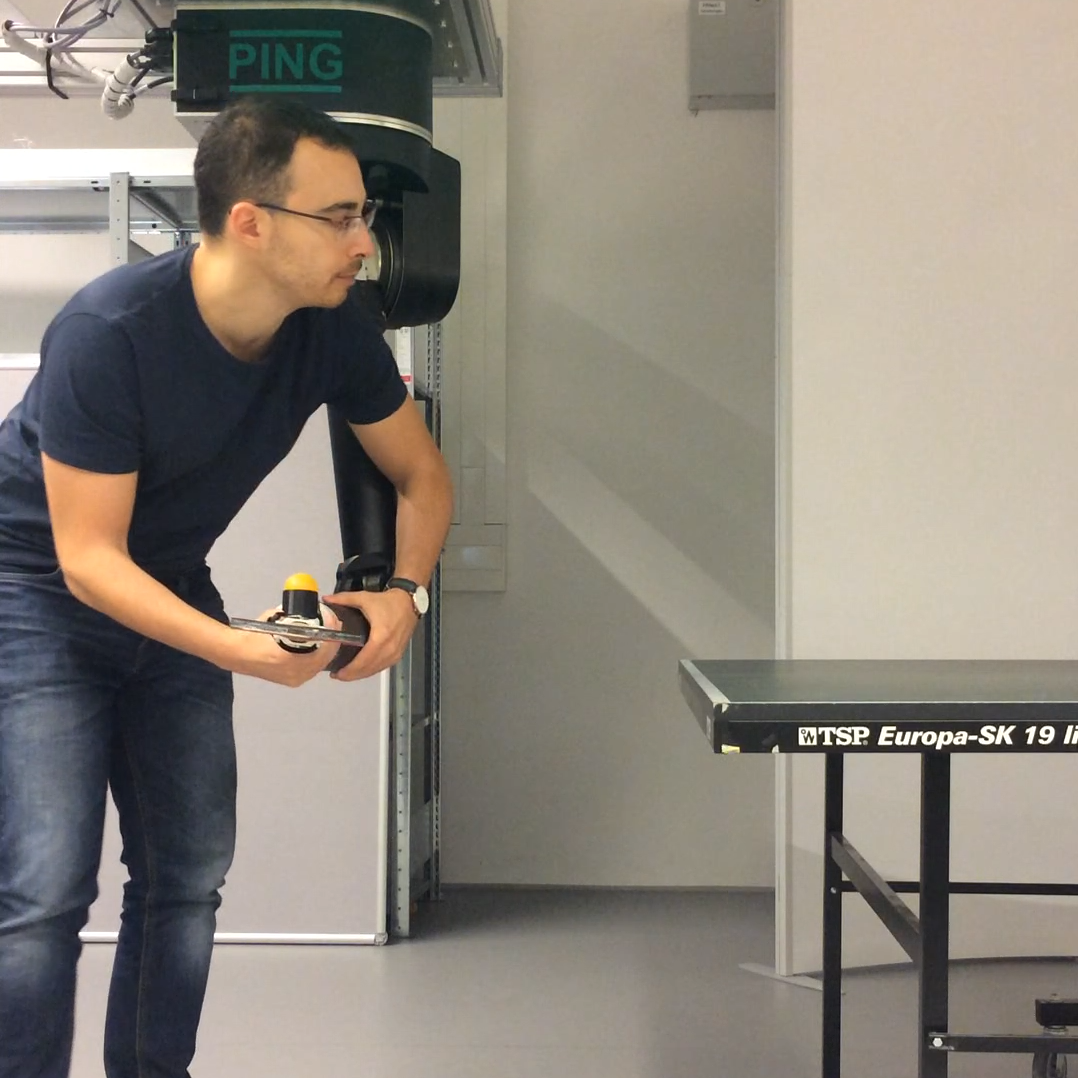}
  \end{minipage}%
  \begin{minipage}{.25\textwidth}
    \includegraphics[scale=0.10]{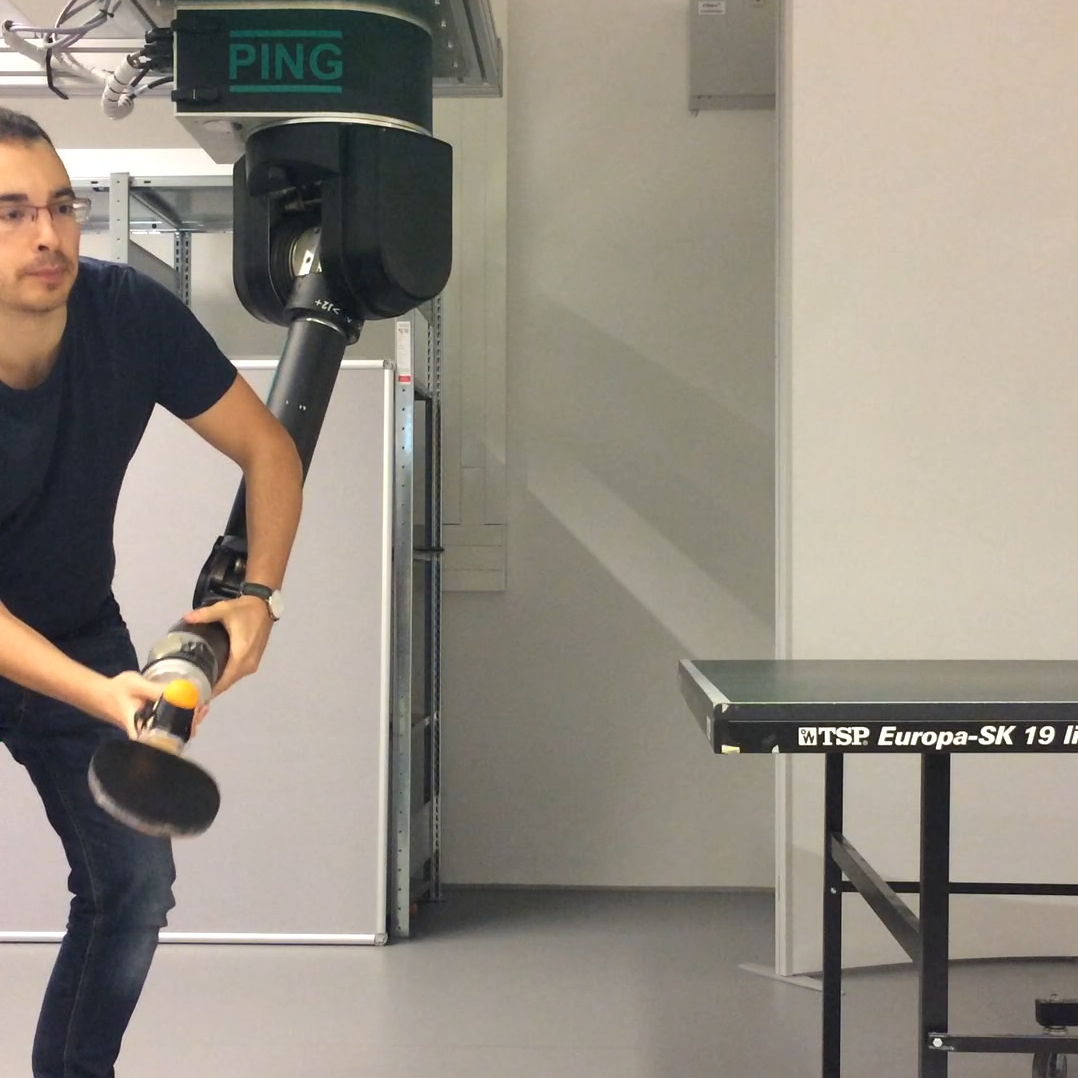}
  \end{minipage}%
  \begin{minipage}{.25\textwidth}
    \includegraphics[scale=0.10]{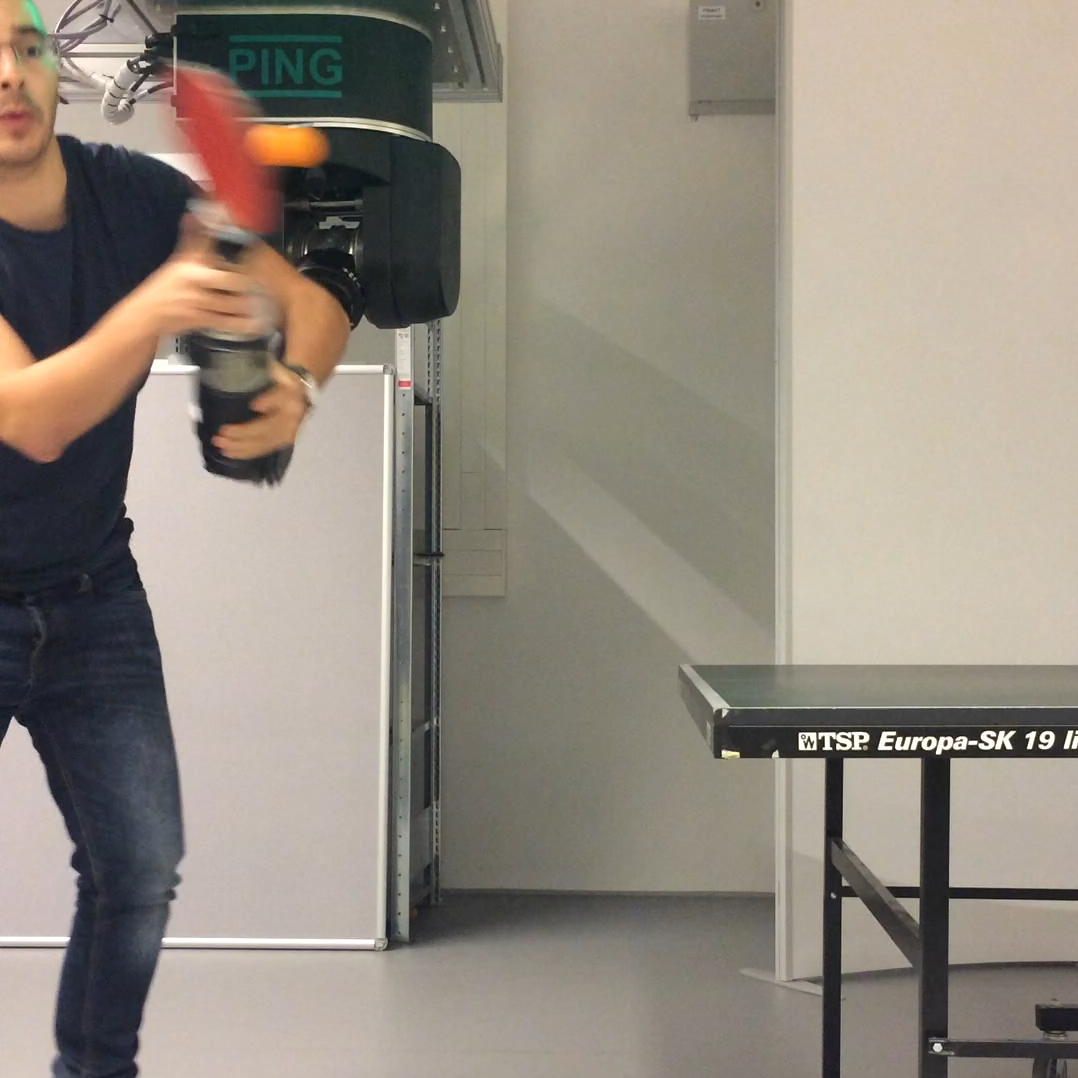}
  \end{minipage}%
  \begin{minipage}{.25\textwidth}
    \includegraphics[scale=0.10]{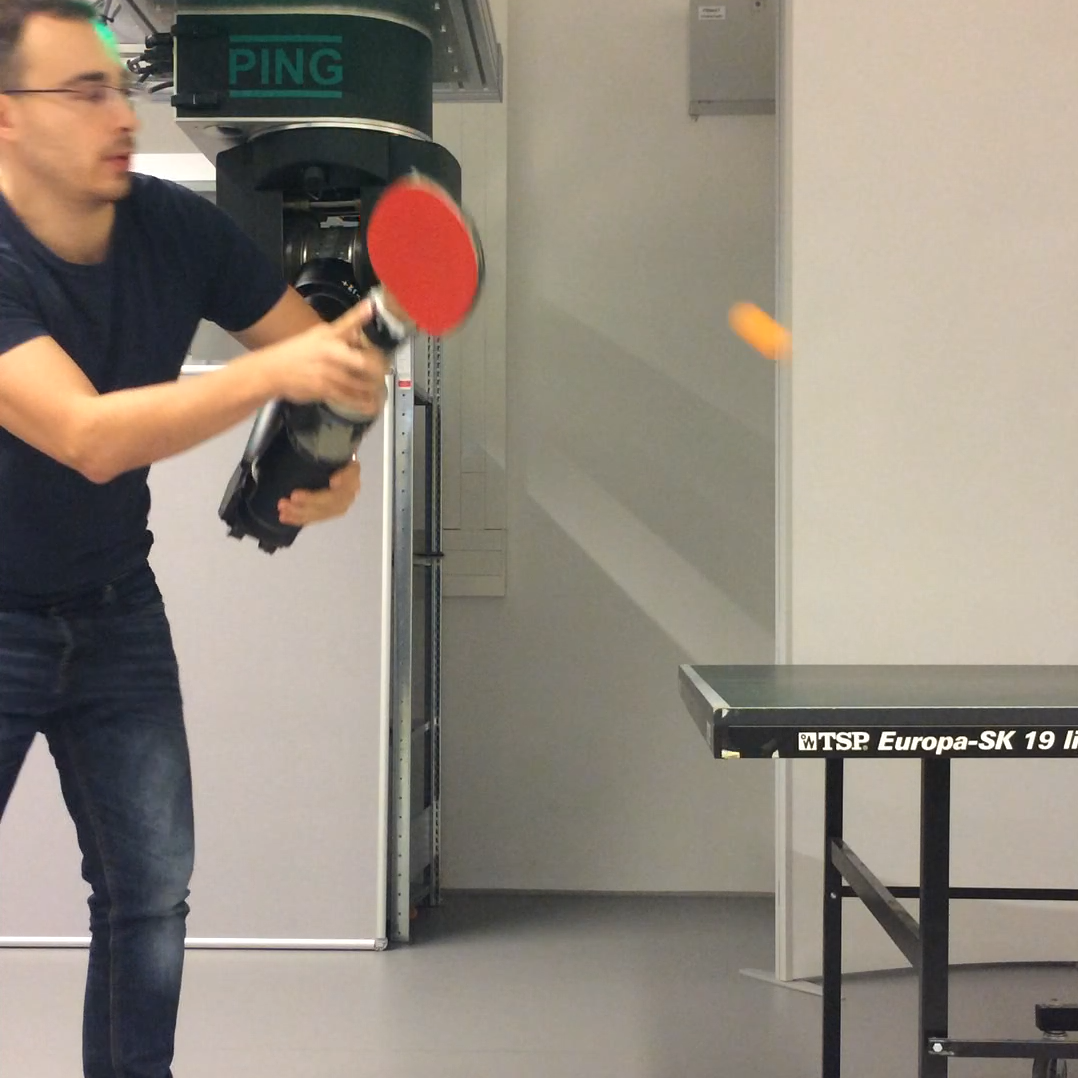}
  \end{minipage}%
  \caption{Our robot table tennis setup with a seven DoF Barrett WAM,
    where we demonstrate, using kinesthetic teach-in, multiple good
    table tennis serve movements while recording the resulting
    joint-space robot trajectories. A metal piece is attached to the
    end effector of the Barrett WAM, which connects to a standard sized
    table tennis racket. An egg-holder on the metal piece holds the
    ball initially before the serve. The demonstrator, after finding a
    good starting posture, starts by swinging the arm, giving the ball
    enough acceleration to propel it away from the robot. The ball is
    then hit in midair by a careful adjustment of the robot
    wrist. The initial posture, the swinging movement of the robot
    shoulder joints and the elbow, and finally the turning of the
    wrist all contribute to the style of the shown movement. Multiple
    demonstrations starting from different initial postures are
    recorded in one session. We compare and evaluate throughout the
    paper different learning from demonstrations approaches using
    these demonstrations. We propose a new iterative optimization
    approach that can learn sparse parameters while adapting the
    features of the movement primitives to the demonstration data.}
  % ready to serve a table tennis ball.
      %Two cameras on the opposite side of the ceiling track the ball
      %continuously at a rate of 180 Hz. 
  \label{robot}
\end{figure*}

In dynamic games like table tennis, we can easily observe humans
taking utmost advantage of their bodies and pushing it to its maximum,
i.e., optimizing their output bearing in mind their kinematic and
dynamic limits. Table tennis serves, for instance, incorporate flicks
(very fast accelerations of the wrist) that are designed to give an
unsuspected spin and motion profile to the ball. Teaching such
movements to the robots in a learning from demonstrations framework
using kinesthetic teach-in, where the robot joint movements are
recorded, suffers in particular from two drawbacks. Firstly, during
the shown movement, as discussed above, the human is unable to move
the shoulder joints of the robot adequately, which could potentially be
used by the robot to great effect. Secondly, the fast movements of the
wrists may not be tracked accurately by the robot, which is the case
for the cable-driven seven degree of freedom (DoF) Barrett WAM arm,
see Figure~\ref{robot}.  % embodied intelligence

In this paper, we explore different learning from demonstrations (LfD)
approaches to compensate for the execution and transfer deficiencies
resulting from the demonstrated serves. The demonstrations are
acquired and the movement primitives are trained in the joint-space of
the robot, using kinesthetic teach-in, where the movements of the
robot are recorded using the joint-level sensors. The initial policy
or the movement template, extracted as a set of movement primitives,
can be thought of as a good initialization for a reinforcement
learning (RL) agent. By capturing the essence of the shown
demonstrations in as few parameters as possible, we simplify and
increase the effectiveness of the skill transfer to the
robot.

Sparsity is achieved in our framework in
joint-space\footnote{Discarding the joint-level information and using
  only the Cartesian coordinates of the resulting movements, in a
  similar attempt to reduce the dimensionality of the robot learning
  problem, necessitates the use of inverse kinematics, running into
  feasibility and additional execution problems that might be
  artificially introduced.} by using a new iterative optimization
approach, where a multi-task Elastic Net regression is alternated with
a nonlinear optimization. The Elastic Net projects the solutions to a
sparse set of features, and during the nonlinear optimization these
features (the basis functions) are adapted to the data in a secondary
optimization. Moreover these features are shared across multiple
demonstrations, increasing the effectiveness of the feature learning
strategy.

The fewer number of learned parameters using our iterative
optimization procedure, compared to more traditional approaches, is
independent of the robot DoF. This is a
desirable property for Reinforcement Learning to adapt the learned
parameters online. Moreover, by using the Elastic Net path, we can
rank the parameters in terms of importance, or effectiveness in
explaining the demonstration data. We perform preliminary experiments
on the Barrett WAM on a table tennis serve to validate the
effectiveness of our new movement primitives.

% certain episodic RL algorithms for a successful serve. As a
% counterpart to learning, we perform experiments where we correct the
% initial serves with online optimization, as in Model Predictive
% Control. We show in the Experiments section in detail the advantages
% and the possible deficiencies of each method.

% Using Model Predictive Control,
% we repeatedly run our optimizer based on new ball positions, making
% the performance more robust to modelling errors and initial
% demonstrations

% We evaluate the efficacy of this new strategy on
% our realistic simulation platform~\cite{Schaal06} that also acts as
% our real-time interface to the robot shown in
% Figure~\ref{robot}.

% Moreover, by running our optimizer repeatedly, we
% can correct for new ball and robot positions, making trajectory
% generation more robust to vision and control errors. 

%%% Local Variables:
%%% mode: latex
%%% TeX-master: "root"
%%% End:

Robot table tennis has, since the nineties, captivated the attention of the robot control and learning communities as a challenging and dynamic task, and research in it has been ongoing ever since. After the pioneering work of Anderson's analytical player~\cite{Anderson88}, there have been various approaches focusing on certain parts of the game, such as simplifications in trajectory generation using a virtual hitting plane~\cite{Ramanantsoa94},~\cite{Muelling11} or learning striking trajectories from demonstrations~\cite{Koc15}. Learning approaches to generate better strikes with Reinforcement Learning (RL) include~\cite{Muelling13}, \cite{Yanlong16}. Recently, \cite{Koc16} has introduced a new trajectory generation framework in table tennis, where they solve a free final-time optimal control problem, generating minimum acceleration striking trajectories. This kinematic optimization approach was extended and evaluated in the real robot table tennis setup in \cite{Koc18}.

The success of this and other similar model-based optimization approaches in dynamic tasks like table tennis heavily depends on the accuracy of the models. In the case of table tennis, an accurate \emph{ball model}~\cite{Nonomura10}, \cite{Koc18} is especially difficult to acquire. The high spin rates make the ball flight difficult to model from first (physical) principles, while the various types of impacts make it also difficult to train machine learning approaches from raw ball position data. For the serve, an additional complication results from the ball take-off phenomena, which is similarly difficult to model or to learn.

% we use this kinematic optimization approach throughout the experiments as a baseline to compare against RL.

Learning from demonstrations (LfD) is a promising framework for
learning various robotic tasks efficiently without using hard-coded
approaches or physical insights to model the specific aspects of each
task. It has been used in many different robot scenarios to great
effect, including robot manipulation and human-robot collaboration
\cite{maeda2017probabilistic}. It was also useful in initializing the
parameters of policy-search RL approaches for robot
learning~\cite{Kober08}. There are, by now, many different frameworks
for LfD, including dynamical system representations such as the
Dynamical Movement Primitives (DMP)~\cite{Ijspeert13}, learning
control Lyapunov-functions~\cite{KhansariZadeh14}, and various other
probabilistic approaches, such as the probabilistic movement
primitives~\cite{Paraschos13} or Gaussian mixture
models~\cite{KhansariZadeh11}. These last two methods can, unlike
DMPs, capture multiple demonstrations in a parametric form, and can
moreover be used to condition on way-points or different targets in
joint or in task space. One particular disadvantage of all of the LfD
approaches introduced above is that the features chosen to regress on
the demonstration data are often manually tuned and the number of
parameters to learn are explicitly specified. We think that fixing the
features and tuning their hyperparameters for particular tasks harm
the generalization and applicability of the movement primitives to
novel scenarios.

The $l_1$-regularized $l_2$-norm regression (from hereon referred to as \emph{Lasso}) is often used in the statistics and machine learning communities as a regression method that can simultaneously also perform automatic feature selection. A detailed introduction and analysis of Lasso can be found in \cite{Hastie01}. Lasso was extended to the \emph{multi-task} case (i.e., multi-output regression with shared features) in \cite{Obozinski06}. Our interest in Lasso lies in the fact that (multi-task) Lasso can perform systematic feature selection while training (multiple) movement primitives, augmenting the applicability of LfD to novel tasks. Moreover, selection and early pruning of features can be used to great effect in RL, possibly reducing the amount of interaction time with the real robot. 

A new incremental procedure to solve ordinary least squares regression as well as Lasso problems was proposed in \cite{Efron04}. This algorithm, called \emph{Least Angle Regression} or \emph{LARS} for short, yields piecewise linear homotopy paths of the regression problem as a function of the $l_1$-regularization term. These paths can be used to rank the features in terms of importance, as will be detailed later. Ranking the features of the trained movement primitives can reduce the curse of dimensionality in RL, decreasing as before the robot interaction time and possibly making the adapted movements also more interpretable to the humans.

The \emph{Elastic Net} imposing additional $l_2$-regularization to Lasso was introduced in \cite{Zou05}, where it was noted that a basic transformation converts the problem to a standard Lasso regression, and this is also valid in the multi-task setting. For the training of movement primitives, especially for dynamic trajectories like the table tennis serves, the \emph{Elastic Net} with its $l_2$-regularization can help to reduce the excessive accelerations throughout the learned movements, making them safer to implement on the robot.

In the next sections, we will detail how the sparse representation-learning of movement primitives can be formulated using the multi-task Elastic Net, coupled with nonlinear optimization on the feature parameters. To the best of our knowledge, the multi-task Elastic Net was not combined before with Radial Basis Functions in a (iterative) nonlinear feature selection and optimization framework. We also think that ranking the learned parameters in terms of importance is a new idea that can benefit the RL community. 
%Elastic Net paths can also be generated with Lars, since the Elastic Net can be converted to a standard Lasso problem.
% mention Reinforcement Learning

%In that paper we have additionally imposed a returning trajectory with a fixed final time, to simplify the real-time trajectory generation process.
%We have suggested a lookup table to load stored feedforward trajectories online based on a fixed starting position.

%%% Local Variables:
%%% mode: latex
%%% TeX-master: "root"
%%% End:

\section{Notation}\label{notation}

The notation that we use throughout the paper is standard: for a robot arm with $n$ degrees of freedom (DoF), the joint configurations are $\joint \in \mathbb{Q} = \{\joint \in \mathbb{R}^n \ | \jointMin \leq \joint \leq \jointMax \}$. The recorded joint positions over a movement is represented as a matrix $\joint(t) \in \mathbb{R}^{\numTimes \times \numMovement}$ of $\numTimes$ rows, with column $i = 1, \ldots, n$ storing the positions throughout the movement corresponding to joint $i$.

Whenever multiple demonstrations are used for learning, i.e., $\joint_{ij}(t)$ is recorded for $i = 1, \ldots, n$ DoF and $j = 1, \ldots, d$ demonstrations, these recordings are stacked to form the $\joints$ matrix. The degrees of freedom are concatenated vertically in this case for a single demonstration, while the columns store the different demonstration data, i.e., $\joint_{ij}(t) \to \joints_{N(i-1) + t/dt, j}$ for a recording of $N$ time points with $dt$ time intervals.

The Frobenius norm of a matrix is the square-root of the sum of its squared elements, $\|\vec{M}\|_{F}^{2}={\sum_i}{\sum_j}m_{ij}^{2}$, whereas the $\|\cdot\|_{21}$ norm used in the multi-task Elastic Net is defined instead as $\|\vec{M}\|_{21}= {\sum_i}\sqrt{{\sum_j} m_{ij}^{2}}$, i.e., $l_2$-norm along the columns (degrees of freedom in our setting) and $l_1$-norm along the rows (time steps). This norm is used to induce sparsity on the features, whose centers are initially located uniformly along the time axis. 

\section{Method}\label{secAlgorithm}

% learn policy instead?
In this section, we discuss how one can acquire a sparse movement pattern from human demonstrations. We present first an algorithm that requires only a single human demonstration, and then present a suitable variant that can be employed for multiple demonstrations. This variant of the algorithm decouples the number of learned parameters from the degrees of freedom of the robot.

\subsection{Learning a sparse representation from a single demonstration}\label{demo}

Given a single demonstration $\joint(\timePoints)$ at the (observed) time points $\timePoints$, we'd like to extract a movement primitive that is sparse. That is, throughout the parametric optimization, we'd like to impose a good fit with as few basis functions as possible, while keeping the accelerations low during the trained movement pattern. Having low accelerations is beneficial both for robot safety as well as improving the tracking (execution) accuracy of the trajectories~\cite{Koc18}. Mathematically, the criterion that we optimize can be written as
% can be easily refined later via (policy search) RL and optimization. 
%
\begin{equation}
  \label{eq:1}
  \min_{\paramsRbf,\paramsRegr}\|\joint(\timePoints) - \basis(\timePoints,\paramsRbf)\paramsRegr\|_F^{2} + \lone\|\paramsRegr\|_{21} + \ltwo\|\basisAcc(\timePoints,\paramsRbf)\paramsRegr\|_F^2,
\end{equation}
\noindent where $\basis(\timePoints,\paramsRbf) \in \mathbb{R}^{N \times p}$ are the evaluations of the basis functions at $\timePoints$, $\paramsRegr \in \mathbb{R}^{p \times n}$ are the (sparse) regression parameters, and $\joint(\timePoints)$ are the joint observations during the shown movement. The nonlinear radial basis functions (RBF) are parameterized by $\paramsRbf \in \mathbb{R}^{p}$.  An $l_2$-penalty is put on the accelerations $\basisAcc(\timePoints,\paramsRbf)\paramsRegr$ of the extracted movement pattern $\basis(\timePoints,\paramsRbf)\paramsRegr$, while a penalty with the $l_1$-norm on the (rows of the) regression parameters $\paramsRegr$ encourages sparsity of the found solutions.

%which can be transformed to a Lasso problem after a suitable transformation, 
% cite Elastic Net paper

This regression problem, for fixed $\paramsRbf$, is known as the multi-task \emph{Elastic Net} in the literature, where the features are shared among the sparse parameters along each degree of freedom. As opposed to the standard (multi-task) Lasso, the $l_2$-norm penalty in the optimization~\eqref{eq:1} penalizing the accelerations throughout the motion, also adds stability to the Lasso solutions~\cite{Zou05}. 

The solution to the weighted Elastic Net problem~\eqref{eq:1} for fixed $\paramsRbf$ can be obtained by transforming the problem to an equivalent (unweighted) Lasso problem, solving it via a convex optimizer (e.g., \emph{coordinate descent} is very effective for Lasso problems), and then transforming the solutions back to the Elastic Net parameters.

We can solve the original problem~\eqref{eq:1} iteratively (as in Expectation-Maximization type of algorithms) by first starting the iteration with a Lasso solution of an overly-parameterized radial basis function regression. At each iteration, the RBF parameters $\paramsRbf_i$ corresponding to the basis functions with nonzero Lasso regression parameters $\paramsRegr_{ij} > 0, \, j \!=\! 1, \ldots, n$ are updated for each $\, i \!=\! 1, \ldots, p$ via nonlinear optimization. The Elastic Net regression is then performed, and the features corresponding to parameters with zero coefficients are removed. These two alternating steps can be continued till convergence, or rather terminated in a fixed number of steps. The iterations converge when the change in function value of the total cost in \eqref{eq:1} is below a certain tolerance $\epsilon$. Depending on the initial solution parameters $\paramsRbf_0$ and $\paramsRegr_0$, the iteration converges to a local minimum.

The full procedure is shown in Algorithm~\ref{alg1} in detail. We call the resulting algorithm \emph{Learning Sparse Demonstration Parameters} or \emph{LSDP} for short. The algorithm alternates between the multi-task Elastic Net (lines $4$ and $10$) and the nonlinear optimizer (BFGS, in line $8$). In between, the zero entries of the regression parameters $\paramsRegr$ and the corresponding columns of $\basis, \basisAcc$ are removed in the Prune step (lines $5$ and $13$). The pruning operation simplifies the optimization in the upcoming iterations, as the removed RBF parameters cannot then be re-elected later. We use the squared exponential kernel to construct our basis functions, i.e., for every $i,j$ we use
\begin{equation}
  \label{eq:2}
\basis_{ij}(t_i) = \exp(-(t_i - \centersRbf_j)^2/(2\widthsRbf_j)) ,
\end{equation}
to form the $(i,j)$'th element of the matrix $\basis$. The data is initially centered (line $2$), i.e., the mean of each joint recording is subtracted from the signal, and the means $\joint_0$ are stored as the intercepts for the particular demonstration. 

For a good performance of the algorithm, i.e., obtaining low residuals with a sparse set and low accelerations, choosing the regularizer weights $\lone$ and $\ltwo$ suitably is crucial. These parameters can be set using cross-validation either before Algorithm~\ref{alg1} or together with the initial regression (line $4$). The regularizers should be scaled down accordingly with the decreasing residual norms (see line $12$), otherwise the algorithm can converge to the empty set for the parameters $\paramsRegr$.

The optimization problem, depending on the parameterization and the features used, can be highly nonconvex, possibly with many local minima. The number of local minima, fortunately, does not seem to pose a problem in terms of residual norm. As long as the initial representation is sufficiently (over) parameterized, most solutions fit well to the demonstration data. For more sparse representations, however, one may choose to restart the training procedure a few times from perturbed initial conditions, especially for the RBF parameters $\paramsRbf$. See the Experiments section for more discussion on the implementation details.
% multiple demonstrations could possibly help refine the basis functions, whose parameters could be stretched or tightened according to the movement speed of the human demonstrator.

The computational complexity of the algorithm overall is dominated by the complexity of the multi-task Elastic Net step (line $10$), where the coordinate descent algorithm is used to solve a Lasso problem (after a transformation in constant time $\mathcal{O}(1)$). The time-complexity of the LARS algorithm to solve Lasso problems is known to be $\mathcal{O}(Np^{2})$ \cite{Efron04}, but coordinate-descent converges often faster, in our experience. One step of Quasi-Newton methods has time-complexity  $\mathcal{O}(p^2)$ (plus the cost for function and gradient evaluations \cite{Nocedal99}), coming from the matrix multiplication operations. Quasi-Newton optimization may, depending on the initialization, require many of these steps, in our case we limit it to $100p$ steps for each iteration of $\alg$.
%\footnote{In practice we see that Quasi-Newton takes much more time than Lasso, on average}. 

%
\begin{algorithm}[t]
%\begin{mdframed}
%\small\sf\centering
\caption{Learning sparse parameters with regression ($\alg$) for a single demonstration}
\label{alg1}
%\begin{minipage}{\linewidth}
\begin{algorithmic}[1]
  \Require $\joint$, $\timePoints$, $\centersRbf$, $\widthsRbf$, $\lone$, $\ltwo$, $\epsilon > 0$
  \State Initialize $\paramsRbf_0 = [\centersRbf,\widthsRbf]$
    \State Center the data, $\joint_0, \joint \leftarrow \code{Center}(\joint)$
   \State Form $\basis, \basisAcc$ using $\paramsRbf_0$ and $\timePoints$
   \State $\paramsRegr_0$ $\leftarrow$ \code{MultiTaskElasticNet}($\basis$, $\basisAcc$, $\joint$, $\lone$, $\ltwo$)
   \State $\paramsRegr_{0}, \paramsRbf_{0}$ $\leftarrow$   \code{Prune}($\paramsRegr_{0}, \paramsRbf_{0}$)
   \State Form $\basis, \basisAcc$ using $\paramsRbf$ and $\timePoints$
   \Repeat $\ k = 1, \ldots, $
   \State $\paramsRbf_{k}$ $\leftarrow$ \code{BFGS}($\basis$, $\basisAcc$, $\paramsRbf_{k-1}$, $\paramsRegr_{k-1}$, $\joint$, $\lone$, $\ltwo$)
   \State Form $\basis,\basisAcc$ using $\paramsRbf_{k}$ and $\timePoints$
   \State $\paramsRegr_{k}$ $\leftarrow$ \code{MultiTaskElasticNet}($\basis$, $\basisAcc$, $\joint$, $\lone$, $\ltwo$)
   \State Calculate residual norm $r_k$, total cost $f_k$ using \eqref{eq:1}
   \State Scale penalties $\lambda_{i} \leftarrow \lambda_i r_{k}^{2}/r_{k-1}^{2}$, $i = 1,2$
   \State $\paramsRegr_{k}, \paramsRbf_{k}$ $\leftarrow$ \code{Prune}($\paramsRegr_{k}, \paramsRbf_{k}$)
   \State Form $\basis, \basisAcc$ using $\paramsRbf$ and $\timePoints$
   \Until{$\|f_k - f_{k-1}\| < \epsilon$}
\end{algorithmic}
%\end{minipage}
%\end{mdframed}
\end{algorithm}

\subsection{Coupling the parameters across dimensions}\label{secCouple}

The algorithm $\alg$ discussed in the previous subsection uses the multi-task Elastic Net to enforce the same basis functions for each degree of freedom (along the columns of $\joint(t)$ and the parameter matrix $\paramsRegr$), while the parameter vectors corresponding to each joint movement are different and optimized independently: the regression parameters are decoupled across the degrees of freedom (DoF) of the robot. In particular, the number of regression parameters grow linearly with the robot DoF, which may be undesirable for applying policy search RL approaches to high dimensional robotic systems especially. 

Furthermore, the algorithm has to be applied for each demonstration separately, i.e., there is no \emph{coupling} or information shared between the demonstrations. In order to enforce rather the features to be shared \emph{across demonstrations} rather than the robot DoFs, we discuss here a variant of the algorithm $\alg$, which we call coupled $\alg$, or $\algCoupled$ for short.

The algorithm $\algCoupled$, shown in Algorithm~\ref{alg2}, requires only a few changes compared to Algorithm~\ref{alg1}. The data is centered for each demonstration to obtain the intercepts $\joints_0$. The algorithm then stacks (lines $1-3$) the dependent regression variables $\joint_i$ and the RBF parameters $\paramsRbf_i$ vertically for each degree of freedom $i = 1, \ldots, n$ to form the matrices $\joints \in \mathbb{R}^{\numTimes\numMovement \times \numDemos}$ and $\basis \in \mathbb{R}^{\numTimes\numMovement \times \numRbf}$. The second time derivative of the data matrix, $\basisAcc$, is stacked as well to form the regression model as in \eqref{eq:1}.

As opposed to $\alg$, in this procedure there are $\numMovement$ times
the number of RBF parameters $\paramsRbf$ to be optimized (line $8$),
as the features are adapted independently for each DoF. The regression
parameters $\paramsRegr$, on the other hand, are coupled across the
DoFs, and their cardinality is reduced by $\numMovement$ times. The
nonlinear optimization computational complexity in this case dominates
that of the multi-task Elastic Net and the net result is roughly a
$\numMovement$ times increase in the computation time between each
iteration of $\algCoupled$.

Note that the parameters for each demonstration are estimated
together, i.e., the columns of the $\paramsRegr$ matrix correspond to
the regression parameters for different demonstrations. One way to
generalize the learned movement primitives to different task
conditions (such as varying initial joint states) would be to
interpolate between these regression parameters. A policy could then be effectively created, whose generalization would be limited by the number and the quality (e.g. variety, success rate) of the demonstrations.

\subsection{Ranking the demonstration parameters}\label{ranking}

\begin{figure}[t]
  \centering
  \setlength{\fwidth}{\columnwidth}
  \setlength{\fheight}{.5\fwidth}
  \input{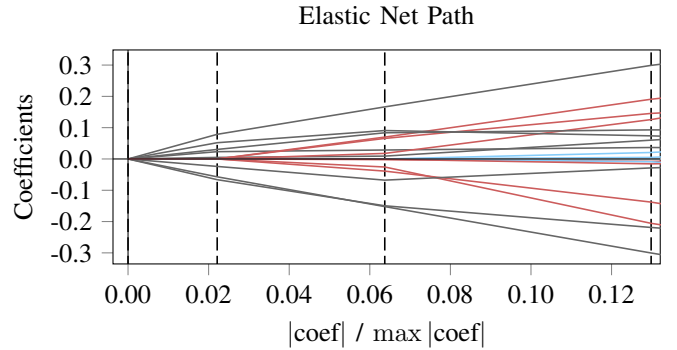}
  \caption{An example Elastic Net path with twenty selected parameters is shown after training Algorithm~\ref{alg2}, $\algCoupled$, with five demonstrations. This \emph{regularization path} can be generated in the final step of the algorithm. As the $l_1$-penalty term $\lambda_1$ of the regression problem \eqref{eq:1} is reduced, the coefficients converge to their (maximal) ordinary least squares values at the right hand side of the plot (not shown). Each dashed line signals an entry of a parameter, and the slope of the coefficients are updated accordingly. The algorithm \emph{LARS}~\cite{Efron04} can be used to generate these piece-wise linear regularization paths. One possible way to use this path is to rank the sparse parameters of the learned movement primitives in terms of statistical importance. For example, in the shown plot, the parameters corresponding to the red lines would be ranked after the other parameters appearing before (black lines). The parameter paths, whose coefficients become nonzero close to each other, are drawn with the same color.}
  \label{elastic_net_path}
\end{figure}

The regression parameters estimated with $\algCoupled$ can also be ranked in terms of statistical significance, i.e., correlation. The Elastic Net \emph{regularization path} of the \emph{LARS} algorithm~\cite{Efron04} traces the evolution of the parameters as the $l_1$-penalty weight $\lambda_1$ of equation \eqref{eq:1} increases. An example regularization path for twenty selected regression parameters $\paramsRegr \in \mathbb{R}^{20}$ are plotted in Figure~\ref{elastic_net_path}. Initially when the regularization is low ($\lambda_1 \approx 0$) on the right side of the Figure, the coefficients are close to their (nonzero) values in ordinary Least Squares. As the regularization term increases, some of these terms drop out, i.e., the coefficients become zero as the path is traced towards the left-hand side of the Figure. The corresponding features can then be eliminated from the regression model, leading not only to a sparse, but also a ranked set of features. 

In the proposed method $\algCoupled$, the \emph{LARS} algorithm
instead of coordinate descent can be used in the final Elastic Net
computation step (line $11$ of Algorithm~\ref{alg2}) to generate the
full regularization path. The addition of the selected movement
primitive parameters can then be traced. An example path for twenty
parameters selected by the Algorithm is plotted in
Figure~\ref{elastic_net_path} against their normalized $l_1$-norm.
These parameters can be ranked according to their evolution, i.e., the
coefficients that early on during the path become nonzero are likely
to signal more causally effective components of the motion. For
example, in the shown plot, the parameters corresponding to the red
lines would be ranked after some of the parameters appearing before
(black lines). More prominent components of the motion can be
identified this way. These movement components could be adapted
earlier with RL strategies, reducing the curse of dimensionality in
high dimensional robot learning problems.
%
%train a ball model offline and an initial policy from demonstrations
% using that initial policy adjust with simple RL method
% whenever the policy looks good run on robot

%
\begin{algorithm}[t!]
%\begin{mdframed}
%\small\sf\centering
\caption{Learning coupled sparse parameters with regression ($\algCoupled$) across multiple demonstrations}
\label{alg2}
%\begin{minipage}{\linewidth}
\begin{algorithmic}[1]
  \Require $\joint_{ij}$, $\timePoints$, $\centersRbf_i$, $\widthsRbf_i$, $\lone$, $\ltwo$, $\epsilon > 0$
  \State Stack $\joint_{ij}$ to form $\joints$, $i \in [1,\numMovement], j \in [1,\numDemos]$
    \State Center the data, $\joints_0, \joints \leftarrow \code{CenterStacked}(\joints)$
   \State Stack $\paramsRbf_0 = [\centersRbf_1,\ldots,\centersRbf_{\numMovement},\widthsRbf_1,\ldots,\widthsRbf_{\numMovement}]$
   \State Stack $\basis, \basisAcc$ using $\paramsRbf_0$ and $\timePoints$ across DoFs
   \State $\paramsRegr_0$ $\leftarrow$ \code{MultiTaskElasticNet}($\basis$, $\basisAcc$, $\joints$, $\lone$, $\ltwo$)
   \State $\paramsRegr_{0}, \paramsRbf_{0}$ $\leftarrow$   \code{PruneStacked}($\paramsRegr_{0}, \paramsRbf_{0}$)
   \State Stack $\basis, \basisAcc$ using $\paramsRbf$ and $\timePoints$ across DoFs
   \Repeat $\ k = 1, \ldots, $
   \State $\paramsRbf_{k}$ $\leftarrow$ \code{BFGS}($\basis$, $\basisAcc$, $\paramsRbf_{k-1}$, $\paramsRegr_{k-1}$, $\joints$, $\lone$, $\ltwo$)
   \State Stack $\basis,\basisAcc$ using $\paramsRbf_{k}$ and $\timePoints$ across DoFs
   \State $\paramsRegr_{k}$ $\leftarrow$ \code{MultiTaskElasticNet}($\basis$, $\basisAcc$, $\joints$, $\lone$, $\ltwo$)
   \State Calculate residual norm $r_k$, total cost $f_k$ using \eqref{eq:1}
   \State Scale penalties $\lambda_{i} \leftarrow \lambda_i r_{k}^{2}/r_{k-1}^{2}$, $i = 1,2$
   \State $\paramsRegr_{k}, \paramsRbf_{k}$ $\leftarrow$ \code{PruneStacked}($\paramsRegr_{k}, \paramsRbf_{k}$)
   \State Stack $\basis, \basisAcc$ using $\paramsRbf$ and $\timePoints$ across DoFs
   \Until{$\|f_k - f_{k-1}\| < \epsilon$}
\end{algorithmic}
%\end{minipage}
%\end{mdframed}
\end{algorithm}

%\subsection{Correcting Movement with Optimization}\label{opt}

% here talk about kinematic optimization details?

%In this section we introduce an algorithm, that solves the optimal control problem introduced above in \eqref{optimProblem2} in reduced form using a $2n+1$ dimensional parametrization. The parametrization has the additional advantage that we can easily include any joint limit constraints that should be enforced throughout the whole motion. The resulting player minimizes sum of squared accelerations during its entire motion while remaining feasible, i.e. satisfying table tennis hitting and landing constraints.
%%% Local Variables:
%%% mode: latex
%%% TeX-master: "root"
%%% End:

\section{Experiments}\label{results}

In this section, we conduct experiments to learn a sparse set of movement primitive parameters using the proposed approaches (see Algorithms \ref{alg1} and \ref{alg2}). The two algorithms are also compared against two competing movement primitive learning methods (DMPs and $l_2$-regularized regression). Finally we present real robot experiments on our table tennis platform where we show that the learned sparse movements nevertheless look similar to the shown demonstrations in style. They can also be implemented safely on the robot.
%In this section, we conduct real robot experiments in our table tennis platform where we compare the performance of policy search (RL) algorithms with a kinematic optimization approach.

%Using Model Predictive Control (MPC), we repeatedly predict the path of the ball after each new filtered ball observation and re-run our algorithm from its current $\joint_0, \dot{\joint}_0$ values provided by the filtered sensor values. 

\subsection{Learning from Demonstrations}

The algorithms $\alg$ and its coupled variant $\algCoupled$, discussed in Section~\ref{secAlgorithm}, are applied here on the demonstrated Barrett WAM serve movements, see Figure~\ref{robot}. From a continuous stream of joint values, recorded at $500$ Hz during a kinesthetic teach-in session, a predetermined number of $d$ movements are selected by detecting the maximum $d$ velocities in joint space and windowing around these points for a fixed duration of one second. We implement the preprocessing as well as the Algorithms in Python, using the \emph{scikit-learn} toolbox for the multi-task Elastic Net and the \emph{scipy} toolbox for the nonlinear optimization (BFGS, see lines $8$ and $9$ in the Algorithms, respectively).

The preprocessed examples using the above procedure result in the joint matrix $\joint(t) \in \mathbb{R}^{500 \times 7}$ for each example demonstration. For the algorithm $\alg$, the initial RBF centers $\centersRbf_0 \in \mathbb{R}^{500}$ are placed at every time point and the RBF widths $\widthsRbf_0$ are set uniformly to $0.1$. The algorithm stretches, prunes and expands the basis functions throughout the optimization to produce a very sparse, nonuniform set of basis functions shared across the seven degrees of freedom (DoF). The columns of the regression parameter matrix $\paramsRegr$, on the other hand, are separate for each DoF.

The Algorithm $\algCoupled$, on the other hand, optimizes $\numMovement$ times more RBF parameters, i.e., $\centersRbf \in \mathbb{R}^{3500}$ and $\widthsRbf \in \mathbb{R}^{3500}$ for the Barrett WAM with $\numMovement = 7$. During the optimization, all of the recorded data from $d$ demonstrations are used together, and the same set of basis function parameters $\paramsRbf = [\centersRbf^{\mathrm{T}}, (\widthsRbf)^{\mathrm{T}}]^{\mathrm{T}}$ are learned across multiple demonstrations. The learned parameters $\centersRbf,
 \widthsRbf, \paramsRegr$, along with the intercepts, are saved after the optimizations to a json file, to be loaded later by the real-time robot controller in $\CC$ during the online experiments.

\begin{figure*}[t]
  \tiny
  % \begin{minipage}{.3\textwidth}
  %   \includegraphics[scale=0.4]{Pictures/Figure_4.pdf}
  % \end{minipage}%
  \begin{minipage}{.5\textwidth}
    \centering
    \includegraphics[width=\textwidth,height=0.21\textheight]{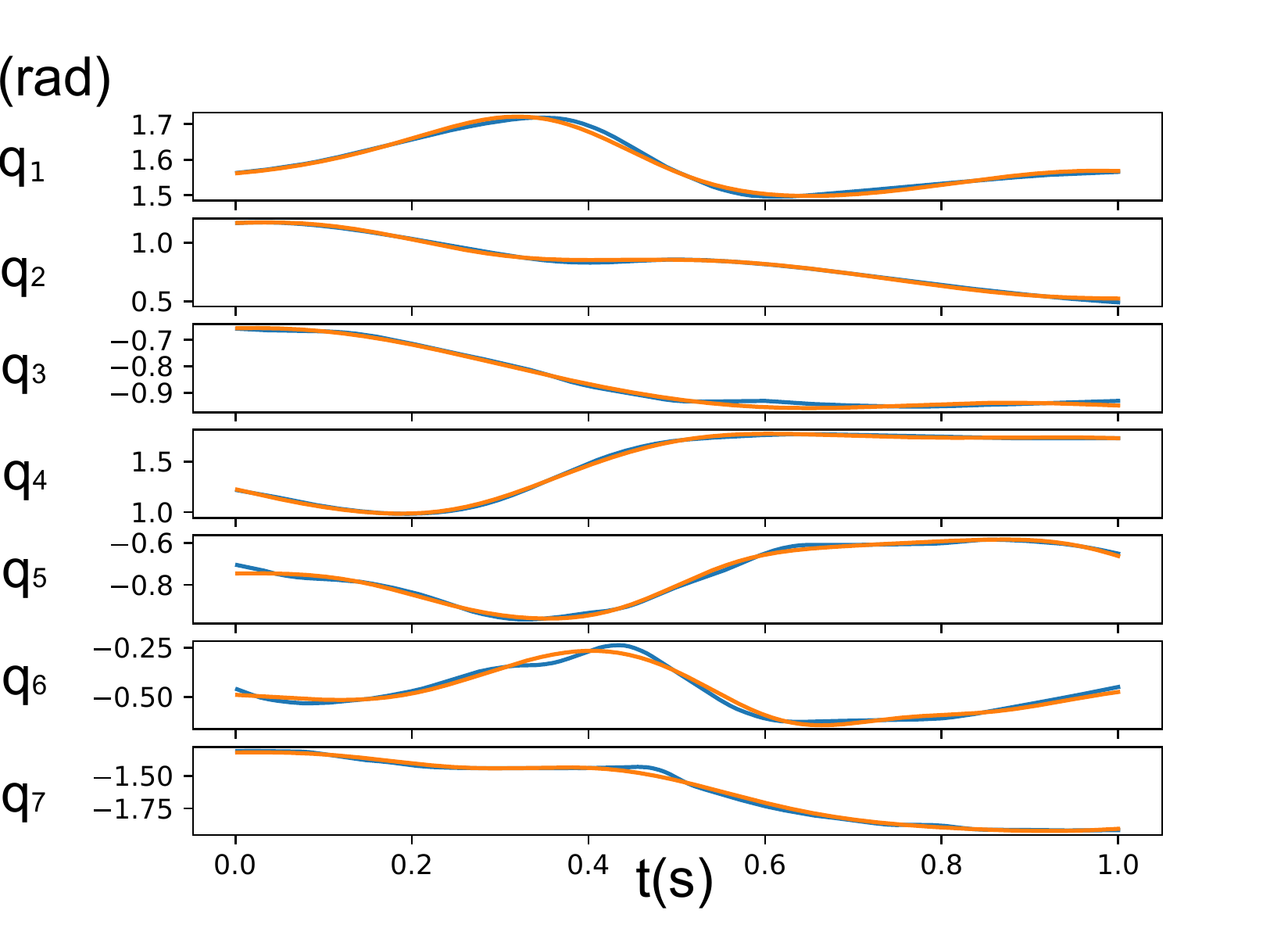}
  \end{minipage}%
  \begin{minipage}{.5\textwidth}
    \centering
    \includegraphics[width=\textwidth,height=0.21\textheight]{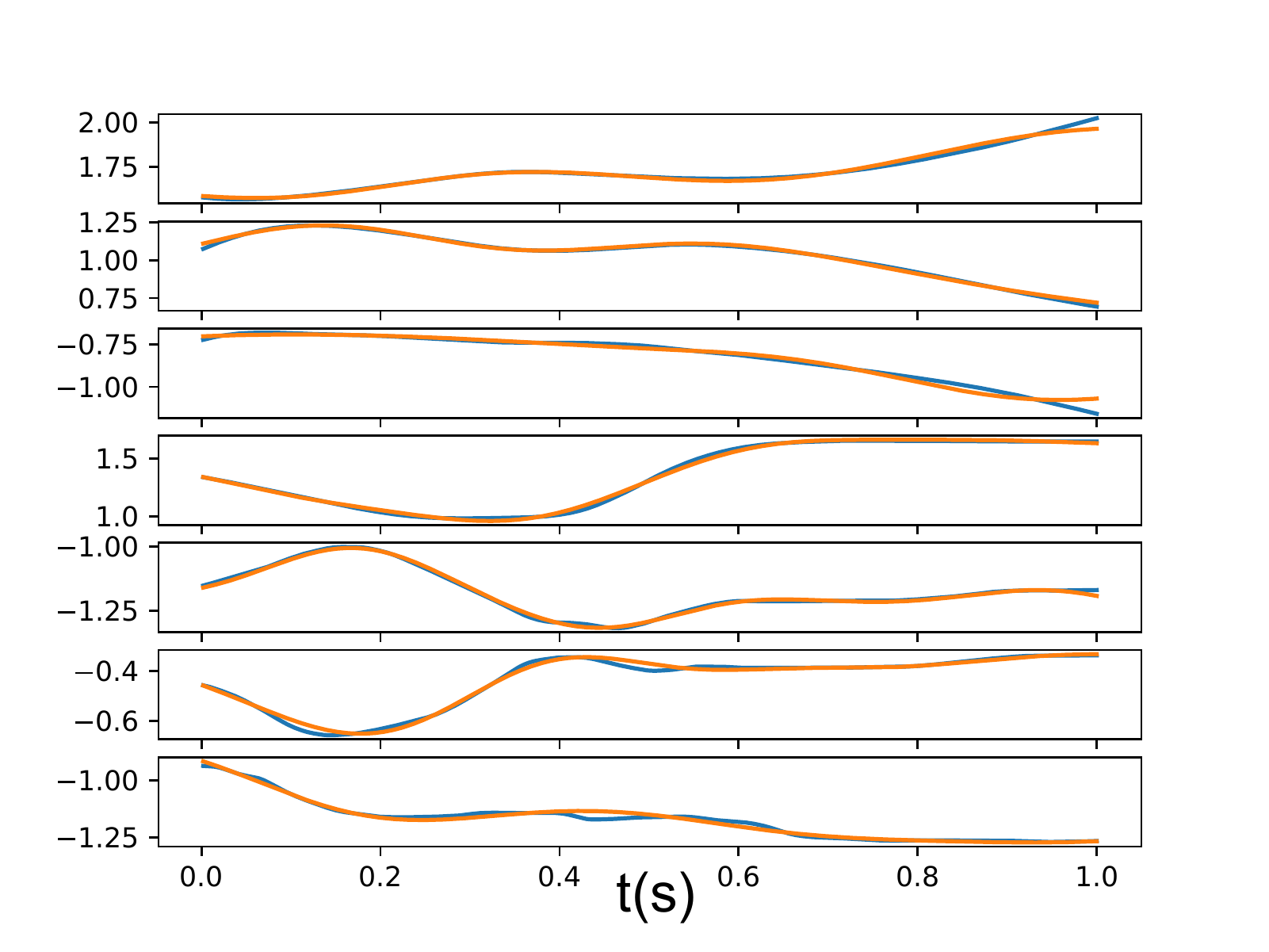}
  \end{minipage}%
  \caption{Two movement primitives learned by the proposed algorithm $\algCoupled$, are plotted in joint space against the recorded demonstrations. The table tennis serve movements, shown in blue, after preprocessing and segmenting the recorded time series are one second long each. The first three rows, $q_1$ through $q_3$, correspond to the shoulder movement in joint space. The fourth row $q_4$ shows the movement of the elbow. Finally, the last three rows ($q_5$ through $q_7$) show the wrist movements in joint space. The trained movement primitives, shown in orange, couple the sparse regression parameters across the degrees of freedom of the robot.}

    %This is a desirable property for a movement primitive serving as a policy in a policy search based reinforcement learning setup, as the number of rollouts needed to improve the expected reward will then be reduced.}
  \label{demo_joint_space}
\end{figure*}
 
Table~\ref{table1} summarizes the results of learning movement primitives from five different demonstrations. The three columns used to compare the different approaches show on average the number of features selected (equivalently, the number of regression parameters with nonzero coefficients), the norm of the second derivatives of the trained movement primitives and the norm of the residuals, respectively. The five demonstration parameters are estimated together in $\algCoupled$, whereas $\alg$ is run separately for each demonstration to obtain the mean and the standard deviations reported in the table. Note that the number of parameters in total used by $\algCoupled \ (37)$ is much lower than the on-average $16.8$ parameters used by $\alg$ for each robot DoF. The residual is slightly higher, this is a result of the parameters being shared across the dimensions. In particular, we have observed that $\algCoupled$ does not fit the last three joints, corresponding to the Barrett WAM wrist, as tightly as $\alg$. This could be because the motion of the wrist is highly varying across the movements and the coupling of the features induced by the algorithm across demonstrations brings these movements closer.

The two proposed algorithms are compared against two baselines in Table~\ref{table1}. The first baseline is the Dynamic Movement Primitives (DMPs) with a fixed number of basis functions. DMPs learn the parameters of an attractor dynamics, i.e., a set of differential equations that converge to a suitably chosen goal state~\cite{Ijspeert13}. A standard regression is performed on the estimated attractor dynamics accelerations. The second baseline is $l_2$-penalized standard regression, with the penalty on the accelerations. During the experiments we used a total of ten basis functions both for the DMPs and for the $l_2$-penalized regression. The basis functions are spread uniformly, as discussed before for the proposed algorithms, around the one second long (preprocessed) demonstrations.

DMPs, as a result of the dynamic constraint of reaching a desired goal position, can incur very high initial accelerations in joint space. Even if the hyperparameters are optimized accordingly to prevent such high accelerations, slight adjustments of initial joint positions can again give rise to high accelerations. The suggestion proposed in \cite{Kober10} to modify the accelerations with the phase can reduce the initial accelerations, but then we have found that the convergence to the goal positions can suffer drastically. The fixed basis function regression does not have this problem, but as in DMPs, optimizes a fixed number of parameters. As shown in Table~\ref{table1}, the number of parameters to fit the demonstrations well is, for both compared methods, on average double the number optimized by $\algCoupled$.

See Figure~\ref{demo_joint_space} for two example regression results. The demonstrated movements are shown in blue and the regression results are shown in orange. The first three rows, $q_1$ through $q_3$, correspond to the shoulder movement in joint space. The fourth row $q_4$ shows the movement of the elbow. Finally, the last three rows ($q_5$ through $q_7$) show the wrist movements in joint space. Although the demonstrated movements are quite different, the training with the sparse set of features can still capture them well.
%see Figure~\ref{robot} for the way the human can hold the robot and show the demonstrations

\begin{table}[b!]
  \sf
  \caption{Comparison of different learning from demonstrations approaches, averaged over five different serve demonstrations}
  \label{table1}
  \begin{tabularx}{0.48\textwidth}{l|l|l|X}
    \toprule
    & \bfseries No. par. \ $(\|\paramsRegr\|_{0})$ & \bfseries Acc. norm & \bfseries Res. norm \\ 
    $\alg$ & $(16.8 \pm 3.25) \times 7$ & $59.04 \pm 7.0$ & $0.59 \pm 0.11$ \\
    $\algCoupled$ & $37$ & $55.98 \pm 11.78$ & $0.73 \pm 0.09$ \\
    \textit{DMPs} & $11 \times 7$ & $621.73 \pm 57.45$ & $0.92 \pm 0.06$ \\
    \scriptsize\textit{$l2$-reg. regr.} & $11 \times 7$ & $215.45 \pm 35.25$ & $2.12 \pm 0.47$
  \end{tabularx}
\end{table}

\begin{figure*}[t]
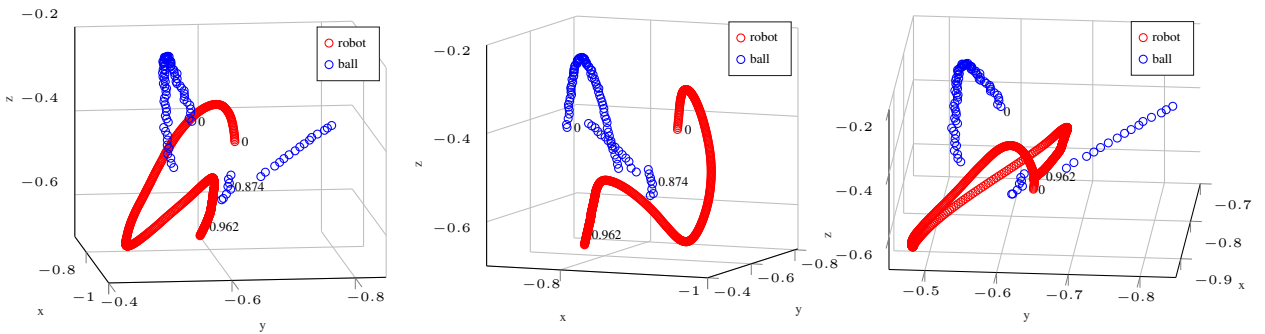

  \centering\tiny
  \begin{minipage}{.3\textwidth}
    \setlength{\fwidth}{0.8\columnwidth}
    \setlength{\fheight}{0.8\fwidth} % golden ratio
    \input{demo_kin.tikz}
  \end{minipage}%
  \begin{minipage}{.3\textwidth}
    \setlength{\fwidth}{0.8\columnwidth}
    \setlength{\fheight}{0.8\fwidth} % golden ratio
    \input{demo_kin2.tikz}
  \end{minipage}%
  \begin{minipage}{.3\textwidth}
    \setlength{\fwidth}{0.8\columnwidth}
    \setlength{\fheight}{0.8\fwidth} % golden ratio
    \input{demo_kin_0.tikz}
  \end{minipage}%

  % \begin{minipage}{0.5\linewidth}
  %   \setlength{\fwidth}{1.0\columnwidth}
  %   \setlength{\fheight}{1.0\fwidth} % golden ratio
  %   \input{../Pictures/demo_joints.tikz}
  % \end{minipage}
  \caption{Three example demonstrations in task space. The initial position of the racket center and the ball in the egg-holder are marked as $0$ in red and blue circles, respectively. The egg-holder is located approximately $14$ cm away from the racket centre. Before the racket stops moving, the ball is already hit, flying towards the table.}
  \label{demo_task_space}
\end{figure*}

Three example demonstrations are plotted in task space in Figure \ref{demo_task_space} along with the recorded ball positions, detected and triangulated from two cameras opposite to the robot. The initial positions of the racket center and the ball in the egg-holder are marked as $0$ in red and blue, respectively. The egg-holder is at a distance of roughly $14$ cm to the racket center. During the movement the ball is hit by the human demonstrator moving the robot arm, and as the demonstrator slows down the motion to a halt, the ball is seen flying towards the table.
% see Figure for an illustration of the recording session with the human demonstrator.

\subsection{Robot Experiments}

Finally, we conduct experiments in our real robot table tennis platform, see Figure~\ref{robot}. Our table tennis playing robot is a seven degree of freedom Barrett WAM arm that is capable of reaching high accelerations and velocities. However it is cable-driven and high accelerations can cause the cables to break easily. A standard size racket is attached to the end-effector via a metal bar. The racket has a radius of roughly $\racketRadius = 7.6$ cm. The table and the table tennis balls are standard sized, balls have a radius of $2$ cm, and the table geometry is roughly $276 \times 152 \times 76$ cm. Throughout the experiments, the Barrett WAM is placed at a distance of about one meter to the end of the table and its base is located $95$ cm above the table. This makes it difficult (but not impossible) for the robot to hit the table. An egg-holder holds the table tennis ball initially, wrapped around the metal bar connecting the end-effector and the racket, see Figure~\ref{robot}.

%To get accurate ball positions during the demonstrations and the robot rollouts, we calibrated the two cameras on the ceiling opposite to the robot with respect to the robot base frame, by collecting pairs of image and robot joints correspondances. The ball detection algorithm is run separately online on every image, running at around 180 frames per second. A linear triangulation algorithm provides then ball position estimates and the resulting estimates of the ball center of mass are then triangulated and filtered with an Extended Kalman Filter (EKG). 
%Alternatively, Cartesian constraints of the table can be incorporated similar to the we way we include the joint limits.

\begin{figure*}[t]
  \centering
  \begin{minipage}{.33\textwidth}
    \includegraphics[scale=0.08]{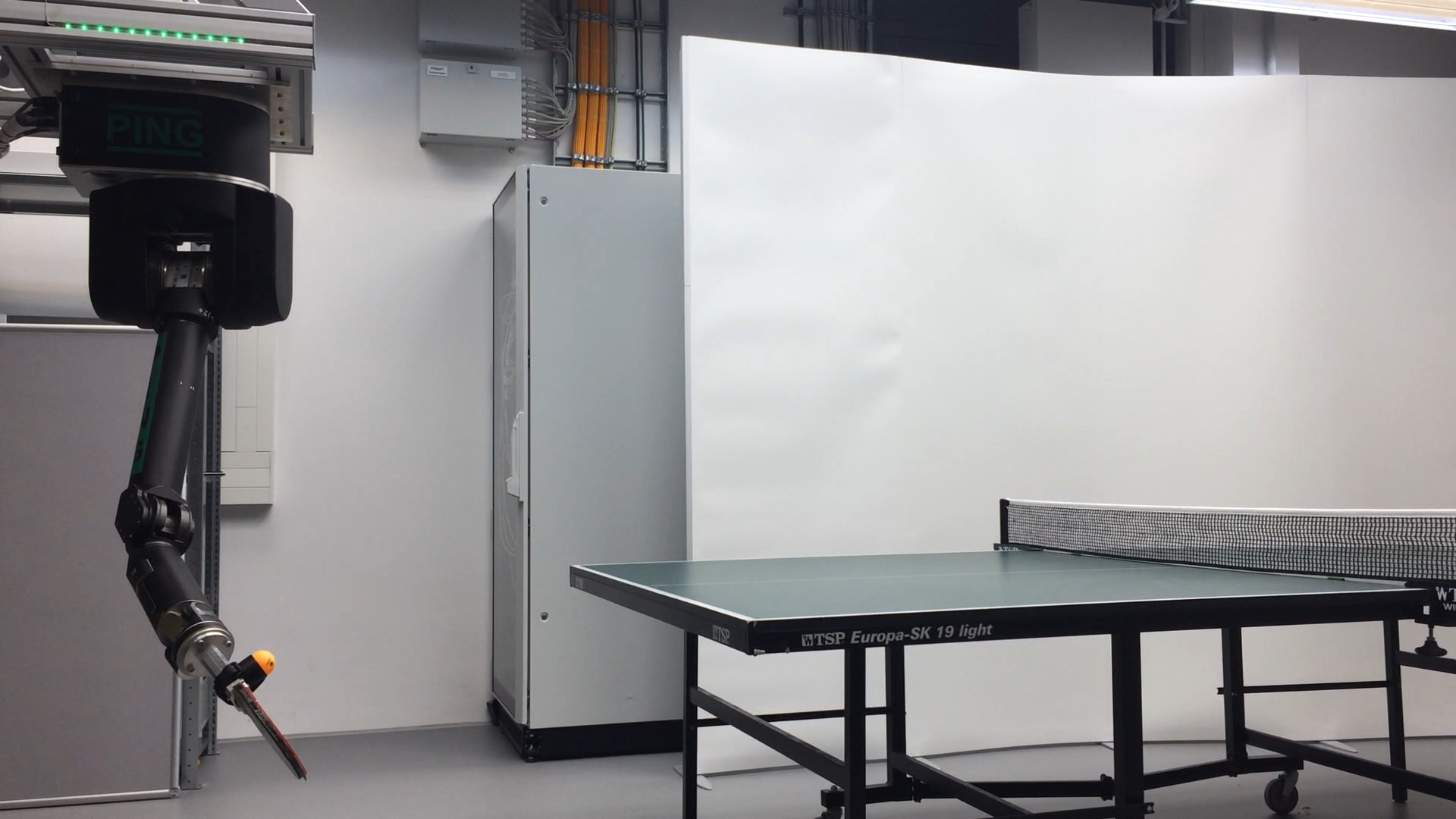}    
  \end{minipage}%
  \begin{minipage}{.33\textwidth}
    \includegraphics[scale=0.08]{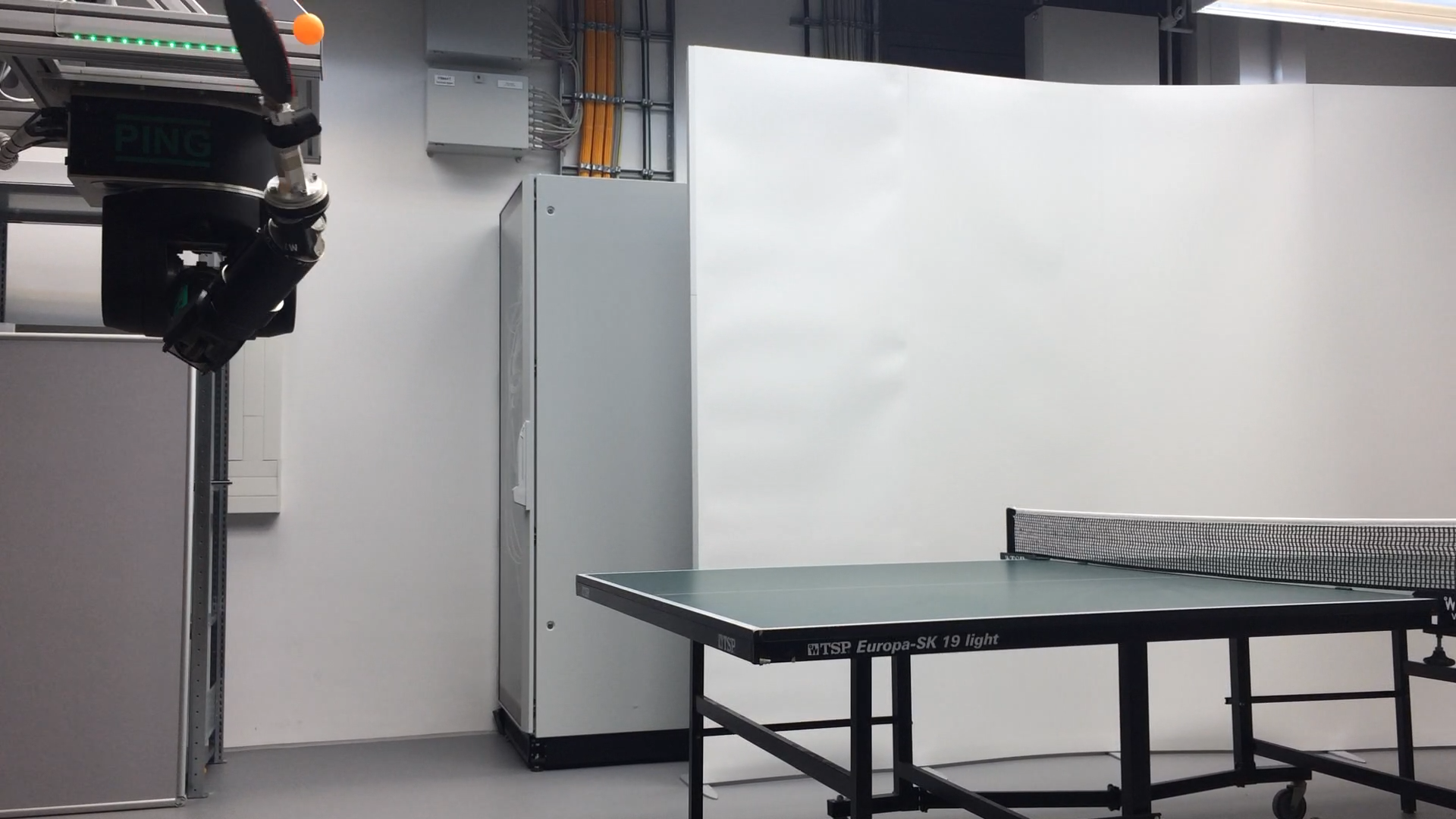}    
  \end{minipage}%
  \begin{minipage}{.33\textwidth}
    \includegraphics[scale=0.08]{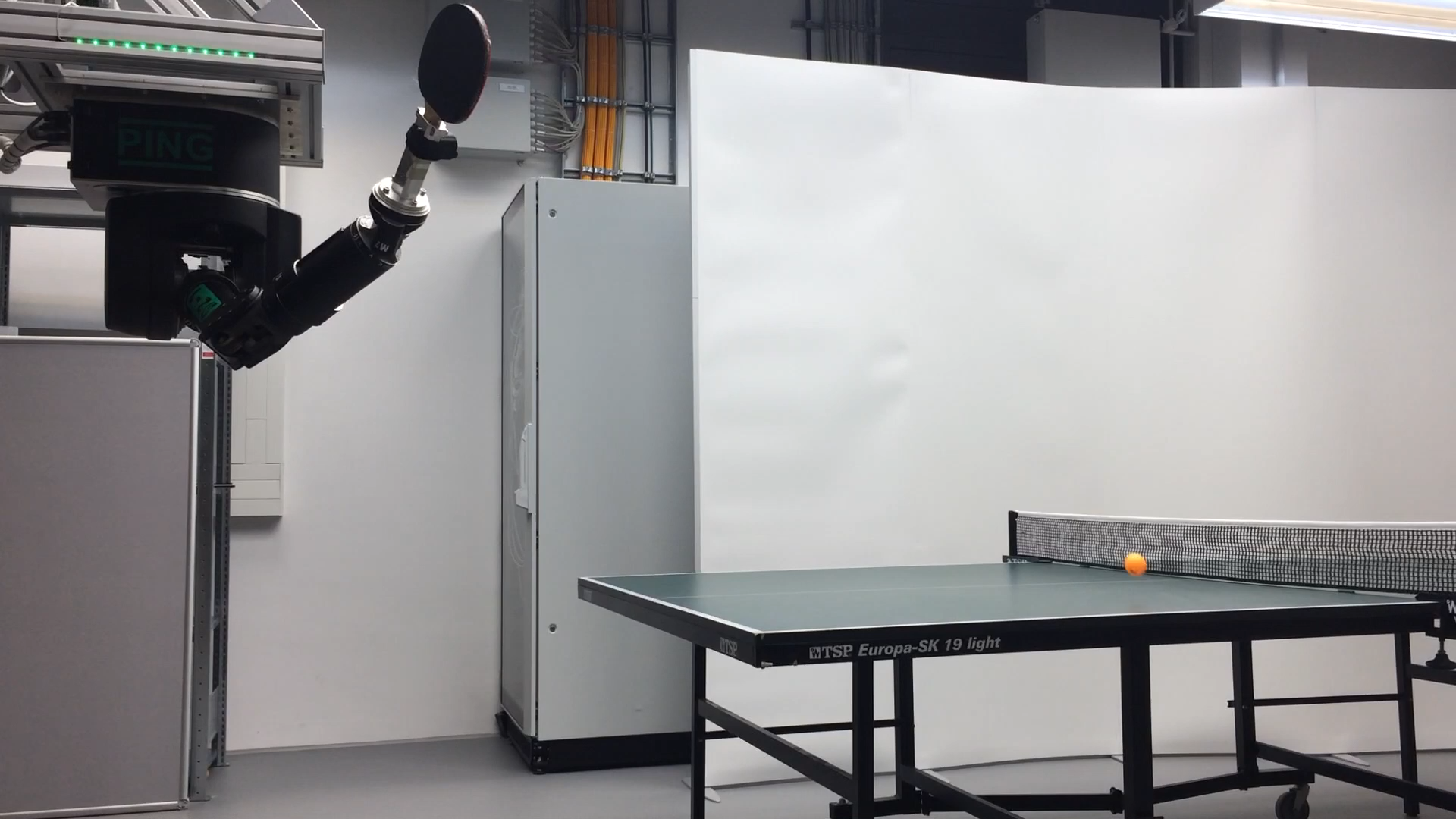}    
  \end{minipage}%
  \caption{A successful rollout during real robot experiments. The ball is initially on top of the egg-holder and during the movement, as a result of the acceleration of the arm, it takes-off from the robot, to be later hit by the racket towards the table. The arm then decelerates towards a safe resting posture.}
  \label{robot_actual_serve}
\end{figure*}

A successful serve in our robot platform is shown in Figure \ref{robot_actual_serve}. The ball is initially placed on top of the egg-holder (approximately $14$ cm away from the racket center along the racket plane). The movements captured by the algorithm $\algCoupled$ are then executed on the robot. During the movements, as a result of the robot's accelerating motion, the ball takes off from the robot arm. The ball is then hit by the robot towards the table. The arm then decelerates towards a resting posture as the ball lands on the robot court, passes the net, and lands again on the opposite side. We notice that the initial accelerating motion and the final wrist movement are critical for a good serve. Without the initial accelerations, the ball has no chance to take-off, and without the final wrist movement (e.g., a quick rotation towards the ball) the ball is not hit well towards the table.

A video showing some demonstrated movements, as well as several actual rollouts on the Barrett WAM is available online: \url{https://youtu.be/vj6jfX_MQmQ}. Note that orchestrating the right movement during the demonstrations can be quite difficult, as moving the shoulder and the elbow joints can feel very awkward depending on the posture. When a good initial posture (both for the robot and the demonstrator) is found, the resulting demonstrations have a higher quality in general. These higher quality demonstrations also have a higher chance of being executed successfully.

Comparing our approach to the DMPs, we notice that the DMPs
immediately start the movement with very large accelerations, these
can be dangerous for the robot and the low-level Barrett WAM
controllers do not support $80 \%$ of the movements. DMPs are good at capturing movements that converge to goal positions (with zero or low velocities), however they are less accurate in capturing the style (e.g., initial movement, final wrist turns) of dynamics movements such as table tennis serves, without manual tuning (the number of basis functions, locations and widths of the basis functions, etc.) for each task. We have seen that $\algCoupled$\footnote{Note that executing the Algorithm $\alg$, trained on each demonstation independently, shows a very similar performance, to that of Algorithm $\algCoupled$ at the moment. However we expect improvements on the learning performance, if Reinforcement Learning is applied on top of the more sparse set of $\algCoupled$ parameters.}
on the contrary, can capture the style of the movement, as shown in Figure $3$ for some of the movements, with a sparse set of basis functions. The generalization capacity of these selected basis functions hinges on the quality and the number of the shown demonstrations.\footnote{If the number and the quality of the demonstrations is not enough, then the selected features and their ranking (using the regularization path) may be spurious, i.e., without any meaningful physical relevance. Increasing the number and the quality (e.g., increased variety of movements, higher success rates) of the movements could remedy such a limitation.}

%We have seen that $\algCoupled$, on the contrary, starts with lower accelerations and catches up later.
% discussion of Dynamic Movement Primitives (DMP) and comparisons
% including Jen's modification

% plot of Algorithm vs. fixed regression for that many parameters
% include 3d plot of balls and robot

% 3 pictures of demonstrations
% 3 pictures of robot actual rollout

% discuss Python -> C++ implementation details
% discuss Calibration

% After the desired trajectories are calculated, high gain PD-control is applied along with an inverse dynamics controller (computed-torque). Our inverse dynamics model is not very precise, however, by running the optimization repeatedly, we're able to compensate and prevent the accumulation of control errors. 

%%% Local Variables:
%%% mode: latex
%%% TeX-master: "root"
%%% End:

\section{Conclusion}\label{end}

In this paper we presented a new learning from demonstrations (LfD) approach to represent and learn table tennis serve movements. The proposed algorithms $\alg$ and $\algCoupled$ learn sparse parameters of the radial basis functions (RBF) from single and multiple demonstrations, respectively. The algorithms employ iterative optimization, alternating between a weighted multi-task Elastic Net regression step that learns sparse parameters given the features and a nonlinear optimization step that adapts the features (more specifically, the widths and centers of the RBFs corresponding to the nonzero regression parameters). The algorithm $\algCoupled$, unlike $\alg$, learns (sparse) parameters that are independent of the robot DoF. This desirable property is achieved by having different basis functions that are adapted across each DoF separately. The multi-task Elastic Net, in this case, forces the joint-dependent features to be shared across multiple demonstrations.

The cost function chosen for the optimization includes the residual of the fit, as well as $l_2$-regularization terms on the accelerations and $l_1$-regularization on the regression coefficients. We compared the performance of the proposed algorithms with Dynamic Movement Primitives (DMPs) and the standard $l_2$-regularized regression, and we evaluated the performance of each on the different components of the chosen cost function (see Table~\ref{table1}). Finally, we discussed the performance of the actual rollouts, using our framework, on the real robot table tennis setup. One can see in the video available online that the style of the movements are preserved while maintaining low accelerations throughout the motion, which is important for the safety of the robot. 
% In this paper we have presented and compared different robot strategies to serve table tennis balls successfully. The serving policies are initialized from human demonstrations and then further adapted with various reinforcement learning (RL) approaches. The parameters of the policies determine the shape and the execution of the robot trajectory.

The sparsity of the parameters, as well as their decoupling from the robot DoF, is a desirable property for policy-search RL approaches that could adapt the regression parameters online based on a suitable reward function. We have presented a way to rank these policy parameters, in the last subsection of Section~\ref{secAlgorithm}, based on how well the parameters explain the (multiple) demonstration recordings. We think that this is a promising research direction to combat the curse of dimensionality in high dimensional robot learning tasks, and we will focus on it more in future experiments.

\bibliographystyle{plain}
\bibliography{ttRef}

\end{document}